\documentstyle[psfig]{llncs}
\begin{document}
\begin{center}
{\large {\bf Geometrical Complexity of Classification Problems}}
\vspace{0.1in}\\
Tin Kam Ho \vspace{0.025in}\\
Bell Laboratories, Lucent Technologies \vspace{0.05in}\\
Nov 20, 2003
\vspace{0.5in}\\
{\bf Abstract}
\vspace{0.1in}\\
\parbox{4.5in}{
Despite encouraging recent progresses in ensemble approaches,
classification methods seem to have reached a plateau in development.
Further advances depend on a better understanding of geometrical and
topological characteristics of point sets in high-dimensional spaces,
the preservation of such characteristics under feature transformations
and sampling processes,  and their interaction with geometrical models
used in classifiers.   We discuss an attempt to measure such
properties from data sets and relate them to classifier accuracies.
}
\end{center}

\section{Introduction}

Advances in ensemble learning have produced a significant rise
in classification accuracy from those achieved when
only monolithic classifiers are known.
However, after the past decade of development, most methods 
seems to have reached maturity, so that no significant improvements
are expected to result from incremental modifications.  Often, for a certain
benchmark problem, one can see many methods in close rivalry, 
producing more or less the same level of accuracy.
Although continuous attempts are being made on
interpreting existing techniques, testing known methods on
new applications, or mix-matching different strategies,
no revolutionary breakthrough appears to be in sight.
It almost seems that a plateau has been reached in 
classification research,  and questions like these begin
to linger in our minds:
1) have we exhausted what can be learned from a given set of data?
2) have we reached the end of classifier development? and
3) what else can be done?

Obviously we have not solved all the classification problem
in the world -- new applications bring new challenges, many of
which are still beyond reach.
But do these require fundamental breakthroughs in classification
research?  Or are these ``merely engineering problems'' that 
will eventually be solved with more machine power, or more likely,
more human power to find better feature extractors and
fine tune the classifier parameters?

To answer these questions we need to know whether there
exists a limit in the knowledge that can be derived from a dataset,
and where this limit lies.  That is, are the classes intrinsically
distinguishable?  And, to what extent are they distinguishable?
These questions are about the intrinsic complexity of a classification
problem, and the match between a classifier's capability to a problem's
intrinsic complexity.  We believe that an understanding of these is
the only way to find out about the current standing of classification
research,  and to obtain insights to guide further developments.
In this lecture we describe our recent efforts along these lines.

\section{Sources of difficulty in classification}

We begin with an analysis of what makes classification difficult.
Difficulties in classification can be traced to three sources: 
1) class ambiguity, 
2) boundary complexity, and
3) sample sparsity and feature space dimensionality.

\subsection*{\bf Class ambiguity}

Class ambiguity refers to the situation when there are cases in a
classification problem whose classes cannot be distinguished using the
given features by {\em any} classification algorithm.
It is often a consequence of the problem formulation.
Classes can be ambiguous for two reasons.  It could be that
the class concepts are poorly defined and intrinsically inseparable.
An example for this is that the shapes of the lower case letter ``l''
and the numeral ``1'' are the same in many fonts
(Figure \addtocounter{figure}{1}\thefigure\addtocounter{figure}{-1}(a)).
Such ambiguity cannot be resolved at the classifier level,
a solution has to involve the application context.  

There is another situation where the classes are well defined, but
the chosen features are not sufficient for indicating such differences
(Figure \addtocounter{figure}{1}\thefigure\addtocounter{figure}{-1}(b)).
Again, there is no remedy at the classifier level.
The samples need to be represented by other features that are more
informative about the classes.
Class ambiguity can occur for only some input cases.
Problems where the classes are ambiguous for at least some cases
are said to have nonzero Bayes error,
which sets a bound on the lowest achievable error rate.

\begin{figure}[h]
\centering
\begin{tabular}{cc}
\begin{tabular}{c}
\psfig{file=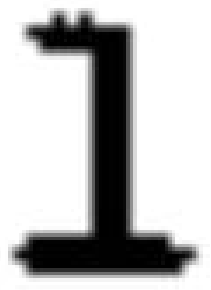,height=0.75in,clip=}\\
\end{tabular} &
\hspace{0.35in}
\begin{tabular}{cccc}
\psfig{file=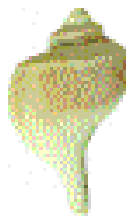,height=0.25in,clip=} &
\psfig{file=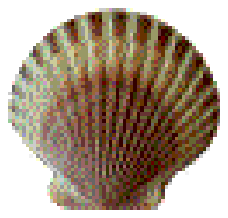,height=0.25in,clip=} &
\psfig{file=figs/scallop.ps,height=0.25in,clip=} &
\psfig{file=figs/whelk.ps,height=0.25in,clip=} \\
\psfig{file=figs/scallop.ps,height=0.25in,clip=} &
\psfig{file=figs/whelk.ps,height=0.25in,clip=} &
\psfig{file=figs/whelk.ps,height=0.25in,clip=} &
\psfig{file=figs/scallop.ps,height=0.25in,clip=} \\
\end{tabular}\\
\parbox{2in}{
(a) The shapes of the lower case letter ``l'' and 
the numeral ``1'' are the same in many fonts. They cannot be
distinguished by shape alone.  Which class a sample belongs to depends
on context.} &
\hspace{0.1in}
\parbox{2in}{
(b) There may be sufficient features for classifying the shells by shape,
but not for classifying by the time of the day when they were collected,
or by which hand they were picked up.}\\
\end{tabular}
\caption{Ambiguous classes due to (a) class definition; (b)
insufficient features.}
\end{figure}

\subsection*{\bf Boundary complexity}

Among the three sources of difficulties,  boundary complexity is
closest to the notion of the intrinsic difficulty of a classification
problem.  Here we choose the class boundary to be the simplest (of
minimum measure in the feature space) decision boundary that minimizes
Bayes error.   With a complete sample, the class 
boundary can be characterized by its Kolmogorov complexity
\cite{kn:kolmogorov65} \cite{kn:li93}.  A class
boundary is complex if it takes a long algorithm to describe, possibly
including a listing of all the points together with their class
labels.
This aspect of difficulty is due to the nature of the problem and is
unrelated to the sampling process.  Also, even if the classes are well
defined, their boundaries may still be complex
(Figure \addtocounter{figure}{1}\thefigure\addtocounter{figure}{-1}).
An example is a random labeling of a set of uniformly distributed
points,  where each point has a definite label, but points of the
same label are scattered over the entire space with no obvious
regularity.  The only way to describe the classes may be an explicit
listing of the positions of the points with the same label.

\begin{figure}[h]
\centering
\begin{tabular}{cccc}
\psfig{file=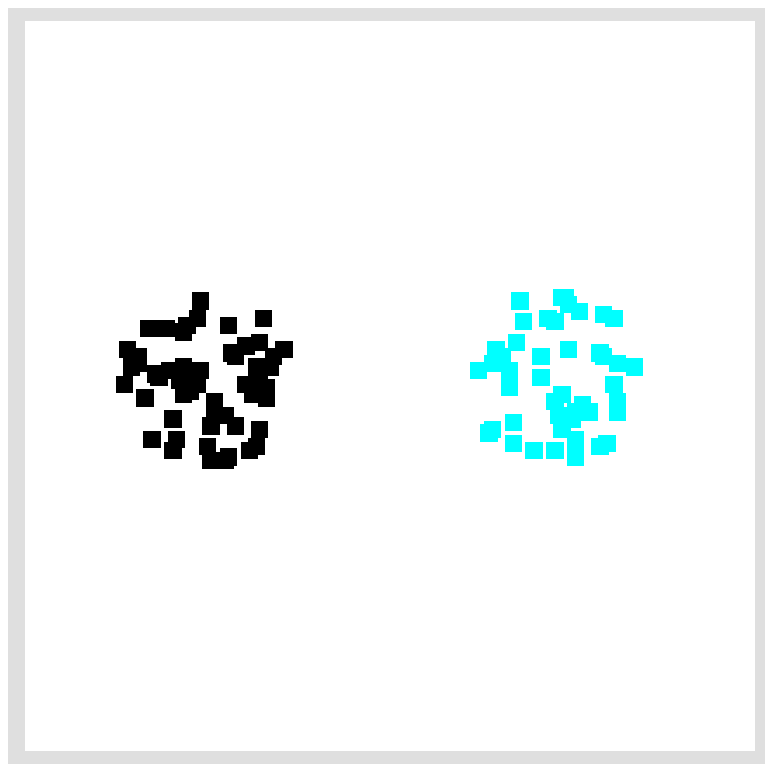,height=0.75in,clip=}&
\hspace{0.2in}
\psfig{file=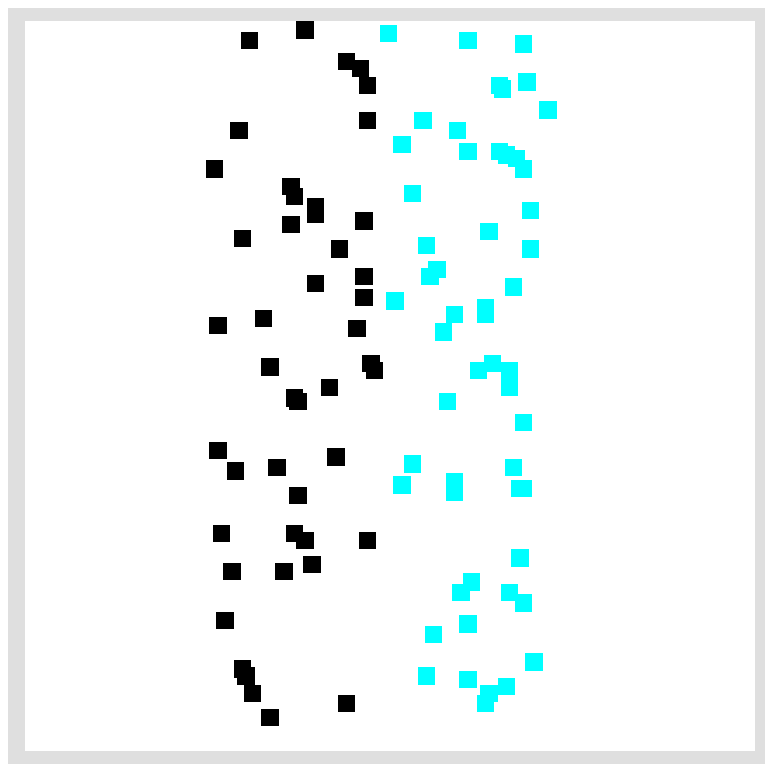,height=0.75in,clip=}&
\hspace{0.2in}
\psfig{file=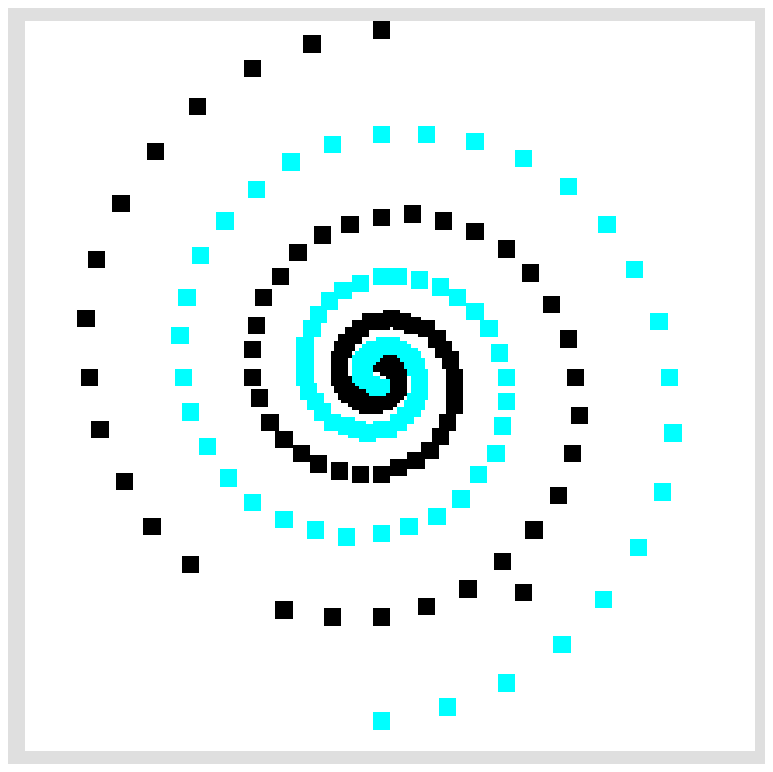,height=0.75in,clip=}&
\hspace{0.2in}
\psfig{file=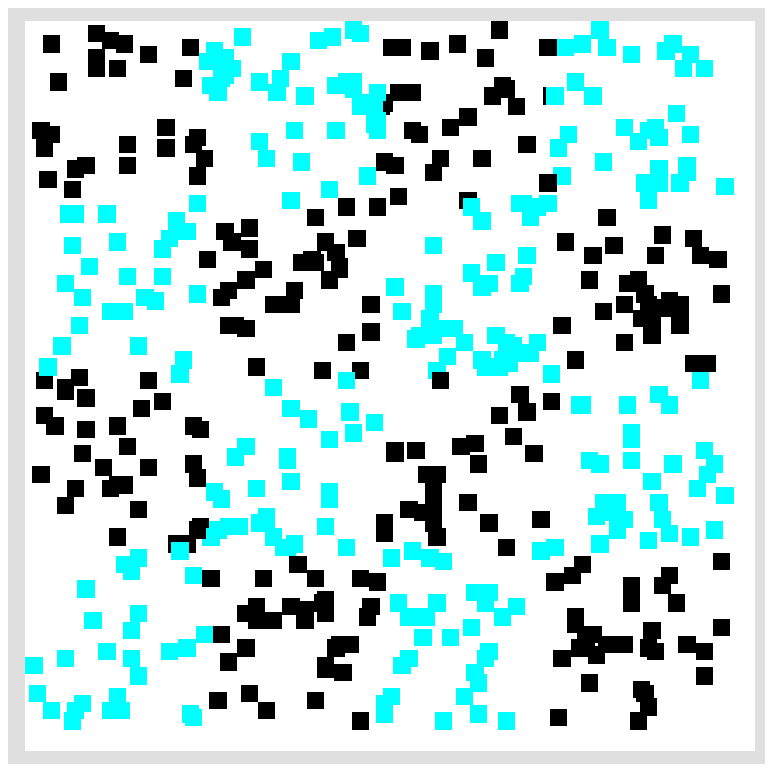,height=0.75in,clip=}\\
(a) & \hspace{0.2in} (b) & \hspace{0.2in} (c) & \hspace{0.2in} (d)\\
\end{tabular}
\caption{Classification problems of different geometrical complexity:
(a) linearly separable problem with wide margins and compact classes;
(b) linearly separable problem with narrow margins and extended
classes; (c) problem with nonlinear class boundary; (d) heavily
interleaved classes following a checker board layout.}
\end{figure}

Kolmogorov complexity describes the absolute amount of information in
a dataset,  and is known to be algorithmically incomputable
\cite{kn:maciejowski79}.  Thus we resort to relative measures that
depend on the chosen descriptors.  Specifically, we can choose a number of
geometrical descriptors that we believe to be relevant in the context
of classification.   We then describe the
regularities and irregularities contained in the dataset in terms of
the chosen geometrical primitives.  We refer to these descriptors as
measures of the {\em geometrical complexity} of a dataset.
This would be sufficient for pattern recognition where most
classifiers can also be characterized by geometrical descriptions of
their decision regions. 

\subsection*{\bf Sample sparsity and feature space dimensionality}

An incomplete or sparse sample adds another layer of difficulty
to a discrimination problem,  since how an unseen point should share
the class labels of the training samples in its vicinity depends
on specific generalization rules.  Without sufficient samples to
constrain a classifier's generalization mechanism,  the decisions on
the unseen samples can be largely arbitrary.
The difficulty is especially severe in high dimensional spaces where
the classifier's decision region, or the generalization rule, can vary
with a large degree of freedom.  The difficulty of working with sparse
samples in high dimensional spaces has been addressed by many other
researchers \cite{kn:devroye88} \cite{kn:raudys91} \cite{kn:vapnik98}.

In practical applications,  often a problem becomes difficult because
of a mixture of boundary complexity and sample sparsity effects.
Sampling density is more critical for an intrinsically complex
problem (e.g. one with many isolated subclasses) than an intrinsically
simple problem (e.g. a linearly separable problem with wide margins),
since longer boundaries need more samples to specify.  If the sample
is too sparse, an intrinsically complex problem may appear deceptively
simple,  like when representative samples are missing from many
isolated subclasses.   However, it can also happen that an
intrinsically simple problem may appear deceptively complex.   An
example is a linearly separable problem that appears to have a
nonlinear boundary when represented by a sparse training set.  Thus,
in lack of a complete sample,  measures of problem complexity
have to be qualified by the representativeness of the training set. 
We will refer to the boundary complexity computed from a fixed training set as
{\em apparent} complexity.

With a given, fixed training set,  there is little one can do
to find out how close the apparent complexity is to the ``true'' complexity.
But this does not prevent one to infer about the true complexity with
some confidence,  if some weak assumptions on the geometry of
the class distributions can be made.
Here we distinguish such assumptions from the more commonly adopted
assumptions on the functional form of class distributions (e.g. Gaussians)
which can be overly strong.   By weak assumptions on class geometry,
we mean those properties such as local compactness of the point sets,
local continuity and piecewise linearity of the boundaries,  all
to be constrained by parameters specifying a small neighborhood.

We believe that even with very conservative assumptions on
the geometrical regularity,  better uses of limited training
samples can be made,  and more useful error estimates can be obtained
than those derived from purely combinatorical arguments emphasizing
the worst cases.  One should be able do these without invoking strong
assumptions on the functional form of the distributions.

\section{Characterization of Geometrical Complexity}

Among the different sources of classification difficulty,  the
geometrical complexity of class boundaries is probably most ready for
detailed investigation.  Thus in this lecture we focus on effective
ways for characterizing the geometrical complexity of classification
problems.

We assume that each problem is represented by a fixed set of training data
consisting of points in a $d$-dimensional real space ${\bf R}^d$, and
that each training point is associated with a class label.
Furthermore, we assume that we have a sparse 
sample, i.e., there are unseen points from the same source
that follow the same (unknown) probability distribution but are
unavailable during classifier design.  The finite and sparse sample
limits our knowledge about the boundary complexity,  thus we are addressing
only the apparent geometrical complexity of a problem based on a
given training set.   We discuss only two-class problems, because most of the
measures we use are defined only for two-class discrimination.
An n-class problem produces a matrix of two-class values for each
chosen measure.   To describe n-class problems,  one needs a way
to summarize such matrices.  There are many ways to do so,
possibly involving cost matrices.   We acknowledge that the
summary by itself is a nontrivial problem.

One natural measure of a problem's difficulty is the error rate of a
chosen classifier.   However, since our eventual goal is to 
study behavior of different classifiers, we want to find other
measures that are less dependent on classifier choices.  Moreover,
measures other than classifier error rates may give hints on how the
errors arise,  which could lead to improvements in classifier
design,  and give guidance on collection of additional samples.

Early in our investigations it became clear that there are multiple
aspects of a problem's complexity that cannot be easily described
by a single known measure.  Furthermore,  while it is easy to 
construct different measures for various characteristics of a dataset, an
arbitrary measure may not necessarily correlate well with any
complexity scale of a reasonable notion.  Such considerations led us
to an evaluation of many different types of measures under a control
study, where each measure is computed for a wide range of problems of
known levels of difficulty.

We constructed a feature (measurement) space for classification
problems,  where each feature dimension is a specific complexity
measure,  and each problem, defined by a labeled training set, is
represented by a point in this space.   Most of the individual
measures came from the literature of both supervised and unsupervised
learning,  with a few others defined by us.  All measures 
are normalized as far as possible for comparability across problems.
The measures we investigated can be divided into several categories:

\begin{enumerate}
\item {\bf Measures of overlaps in feaure values from different classes.}
These measures focus on the effectiveness of a single feature
dimension in separating the classes,  or the composite effects 
of a number of dimensions.   They examine the range and spread of
values in the dataset w.r.t. each feature, and check for overlaps
among different classes
(Table \addtocounter{table}{+1}\thetable\addtocounter{table}{-1}).

\item {\bf Measures of separability of classes.}
These measures evaluate to what extent two classes are separable by
examining the existence and shape of the class boundary.
The contributions of different feature dimensions are summarized in
a distance metric rather than evaluated individually
(Table \addtocounter{table}{+2}\thetable\addtocounter{table}{-2}).

\item {\bf Measures of geometry, topology, and density of manifolds.}
These measures give indirect characterizations of class separability.
They assume that a point class is made up of single or multiple
manifolds which form the support of the probability distribution
of the given class.   The shape, position, and interconnectedness of
these manifolds give hints on how well two classes are separated,
but they do not describe separability by design
(Table \addtocounter{table}{+3}\thetable\addtocounter{table}{-3}).
\end{enumerate}

\begin{table}[ht]
\center
\begin{tabular}{|l|l|l|} \hline
F1 & \multicolumn{2}{l|}{maximum Fisher's discriminant ratio} \\
\cline{2-3}
& \hspace{0.0125in}
\parbox{2.75in}{
Fisher's discriminant ratio for one feature dimension is defined as:
\vspace{0.05in}\\
$f ~=~ \frac{(\mu_1 - \mu_2)^2}{{\sigma_1}^2 + {\sigma_2}^2}$ 
\vspace{0.05in}\\
where $\mu_1$, $\mu_2$, ${\sigma_1}^2$, ${\sigma_2}^2$ are
the means and variances of the two classes respectively, in that
feature dimension.
We compute $f$ for each feature and take the maximum as measure F1.
} 
\hspace{0.0125in} & \hspace{0.0125in}
\parbox{1.5in}{
For a multi-dimensional problem,  not necessarily all features have to
contribute to class discrimination.  As long as there exists one
discriminating feature,  the problem is easy.  Therefore
the maximum $f$ over all feature dimensions is the one most relevant
to class separability.}  
\hspace{0.0125in}
\\ \hline
F2 & \multicolumn{2}{l|}{volume of overlap region} \\
\cline{2-3}
& 
\hspace{0.0125in}
\parbox{2.75in}{
Let the maximum and minimum values of each
feature $f_i$ in class $c_j$ be $max(f_i, c_j)$ and $min(f_i, c_j)$, 
then the overlap measure F2 is defined to be
\vspace{0.05in}\\
$ {\rm F2} = \prod_i \frac{{\rm MINMAX}_i - {\rm MAXMIN}_i}
{{\rm MAXMAX}_i - {\rm MINMIN}_i}$
\vspace{0.05in}\\
where $i = 1,...,d$ for a $d$-dimensional problem, and
\vspace{0.05in}\\
{\small
$ {\rm MINMAX}_i = {\rm MIN} (max(f_i,c_1), max(f_i,c_2)) $\\
$ {\rm MAXMIN}_i = {\rm MAX} (min(f_i,c_1), min(f_i,c_2)) $\\
$ {\rm MAXMAX}_i = {\rm MAX} (max(f_i,c_1), max(f_i,c_2)) $\\
$ {\rm MINMIN}_i = {\rm MIN} (min(f_i,c_1), min(f_i,c_2)) $
}
}
\hspace{0.0125in} & \hspace{0.0125in}
\parbox{1.5in}{
F2 measures the amount of overlap of the bounding boxes of two
classes.  It is a product of the per-feature ratio of the size
of the overlapping region over the size of total occupied region by
the two classes.   The volume is zero as long as there is at
least one dimension in which the value ranges of the two classes are distinct.}
\hspace{0.0125in}
\\ \hline
F3 & \multicolumn{2}{l|}{maximum (individual) feature efficiency} \\
\cline{2-3}
\hspace{0.0125in} & \hspace{0.0125in}
\parbox{2.75in}{
In a procedure that progressively removes unambiguous points falling 
outside the overlapping region in each chosen dimension
\cite{kn:ho98b}, the {\em efficiency} of each feature is defined as
the fraction of all remaining points separable by that feature.
To represent the contribution of the feature most useful in
this sense,  we use the {\em maximum feature efficiency} (largest
fraction of points distinguishable with only one feature) as a measure
(F3).
}
\hspace{0.0125in} & \hspace{0.0125in}
\parbox{1.5in}{
This measure considers only separating hyperplanes perpendicular
to the feature axes.  Therefore, even for a linearly separable problem,
F3 may be less than 1 if the optimal separating hyperplane happens to
be oblique.
}
\hspace{0.0125in} \\ \hline
\end{tabular}
\caption{Measures of overlaps in feaure values from different classes.}
\end{table}

\begin{table}[ht]
\center
\begin{tabular}{|l|l|l|} \hline
L1 & \multicolumn{2}{l|}{minimized sum of error distance by linear
programming (LP)}\\ 
\cline{2-3}
& \hspace{0.0125in}
\parbox{2.75in}{
Linear classifiers can be obtained by a linear programming formulation
proposed by Smith \cite{kn:smith68} that minimizes the sum of distances of
error points to the separating hyperplane (subtracting a constant margin):
\begin{center}
\begin{tabular}{r l}
${\rm minimize~~} $ &  ${\bf a}^t {\bf t}$\\

${\rm subject~to}$ & ${\bf Z}^t {\bf w} + {\bf t} \geq {\bf b}$\\
		   & ${\bf t} \geq {\bf 0}$\\
\end{tabular}
\end{center}

\noindent where ${\bf a}$, ${\bf b}$ are arbitrary constant vectors
(both chosen to be ${\bf 1}$), ${\bf w}$ is the weight vector,  ${\bf
t}$ is an error vector,  and  ${\bf Z}$ is a matrix where each column
${\bf z}$ is defined on an input vector ${\bf x}$ (augmented by adding
one dimension with a constant value 1) and its class $c$ (with value
$c_1$ or $c_2$) as follows: 

\[ \left\{\begin{array}{ll}
{\bf z} =+ {\bf x} & \mbox{if $c= c_1$} \\
{\bf z} = - {\bf x} & \mbox{if $c= c_2$.}
\end{array}
\right. \]

The value of the objective function in this formulation is used as a
measure (L1). 
} 
\hspace{0.0125in} & \hspace{0.0125in}
\parbox{1.5in}{
The measure is zero for a linearly separable problem.  Notice that
this measure can be heavily affected by outliers that happen to be on
the wrong side of the optimal hyperplane.  We normalize this measure
by the number of points in the problem and also by the length of the
diagonal of the hyperrectangular region enclosing all training points in the
feature space.  }
\hspace{0.0125in}
\\ \hline
L2 & \multicolumn{2}{l|}{error rate of linear classifier by LP}\\
\cline{2-3}
& \hspace{0.0125in}
\parbox{2.75in}{
This measure is the error rate of the linear classifier
defined for L1, measured with the training set.}
\hspace{0.0125in} & \hspace{0.0125in}
\parbox{1.5in}{
With a small training set this may be a severe underestimate
of the true error rate.
} \hspace{0.0125in} \\ \hline
N1 & \multicolumn{2}{l|}{fraction of points on boundary (MST method)}\\
\cline{2-3}
& \hspace{0.0125in}
\parbox{2.75in}{
This method constructs a class-blind MST over the entire dataset,
and counts the number of points incident to an edge going across the
opposite classes.   The fraction of such points over all points in the
dataset is used as a measure.} 
\hspace{0.0125in} & \hspace{0.0125in}
\parbox{1.5in}{
For two heavily interleaved classes, a majority of points are located
next to the class boundary.  However, the same can be true for a linearly
separable problem with margins narrower than the distances between
points of the same class. 
} \hspace{0.0125in} \\ \hline
N2 & \multicolumn{2}{l|}{ratio of average intra/inter class NN distance}\\
\cline{2-3}
& \hspace{0.0125in}
\parbox{2.75in}{
We first compute the Euclidean distance from each point to its nearest
neighbor within the class, and also to its nearest neighbor outside
the class.  We then take the average (over all points) of all the
distances to intra-class nearest neighbors,  and the average of 
all the distances to inter-class nearest neighbors.  The ratio of 
the two averages is used as a measure.
}
\hspace{0.0125in} & \hspace{0.0125in}
\parbox{1.5in}{
This compares the within-class spread to the size of the gap
between classes.  It is sensitive to the classes of the
closest neighbors to a point, and also to the difference in magnitude of the
between-class distances and that of the with-class distances.
} \hspace{0.0125in} \\ \hline
N3 & \multicolumn{2}{l|}{error rate of 1NN classifier} \\
\cline{2-3}
& \hspace{0.0125in}
\parbox{2.75in}{
This is simply the error rate of a 
nearest-neighbor classifier measured with the training set.
} 
\hspace{0.0125in} & \hspace{0.0125in}
\parbox{1.5in}{
The error rate is estimated by the leave-one-out method.
} \hspace{0.0125in} \\ \hline
\end{tabular}
\caption{Measures of class separability.}
\end{table}

\begin{table}[ht]
\center
\begin{tabular}{|l|l|l|} \hline
L3 & \multicolumn{2}{l|}{nonlinearity of linear classifier by LP} \\
\cline{2-3}
& \hspace{0.0125in}
\parbox{2.75in}{
Hoekstra and Duin \cite{kn:hoekstra96} proposed a measure for the
{\em nonlinearity} of a classifier w.r.t. to a given dataset.
Given a training set,  the method first creates a test set by
linear interpolation (with random coefficients) between randomly drawn
pairs of points from the same class.   Then the error rate of the
classifier (trained by the given training set) on this test set is
measured.  Here we use such a nonlinearity measure for 
the linear classifier defined for L1.
} 
\hspace{0.0125in} & \hspace{0.0125in}
\parbox{1.5in}{
This measure is sensitive to the smoothness of the classifier's
decision boundary as well as the overlap of the convex hulls of the classes.
For linear classifiers and linearly separable problems,  it measures
the alignment of the decision surface with the class boundary.  It carries
the effects of the training procedure in addition to those of
the class separation. 
} \hspace{0.0125in} \\ \hline
N4 & \multicolumn{2}{l|}{nonlinearity of 1NN classifier} \\
\cline{2-3}
& \hspace{0.0125in}
\parbox{2.75in}{
This is the nonlinearity measure, as defined for L3, calculated for 
a nearest neighbor classifier.} 
\hspace{0.0125in} & \hspace{0.0125in}
\parbox{1.5in}{
This measure is for the alignment of the nearest-neighbor boundary with the
shape of the gap or overlap between the convex hulls of the classes.
} \hspace{0.0125in} \\ \hline
T1 & \multicolumn{2}{l|}{fraction of points with associated adherence
subsets retained}\\
\cline{2-3}
& \hspace{0.0125in}
\parbox{2.75in}{
This measure originated from a work on describing shapes of class
manifolds based on a notion of {\em adherence subsets} in pretopology
\cite{kn:lebour96}.  Simply speaking,  it counts the number of balls
needed to cover each class, where each ball is centered at a training
point and grown to the maximal size before it touches another class.
Redundant balls lying completely in the interior of other balls are removed.
We normalize the count by the total number of points. 
}
\hspace{0.0125in} & \hspace{0.0125in}
\parbox{1.5in}{
A list of such balls is a composite description of the shape of the
classes.  The number and size of the balls
indicate how much the points tend to cluster in hyperspheres or
spread into elongated structures.  In a problem where each point is
closer to points of the other class than points of its own class,
each point is covered by a distinctive ball of a small size, resulting
in a high value of the measure.  
} \hspace{0.0125in} \\ \hline
T2 & \multicolumn{2}{l|}{average number of points per dimension}\\
\cline{2-3}
& \hspace{0.0125in}
\parbox{2.75in}{
This is a simple ratio of the number of points in the dataset over the
number of feature dimensions.}
\hspace{0.0125in} & \hspace{0.0125in}
\parbox{1.5in}{
This measure is included mostly for connection with prior studies on
sample sizes.   Since the volume of a region scales exponentially
with the number of dimensions,  a linear ratio between the two is not
a good measure of sampling density.
} \hspace{0.0125in} \\ \hline
\end{tabular}
\caption{Measures of geometry, topology, and density of manifolds.}
\end{table}

Many of these measures have been used before, in isolation, to
characterize classification problems.  But there have been little 
serious studies on their effectiveness.  Some are known to be good
only for certain types of datasets.  For instance, Fisher's
discriminant ratio is good for indicating the separation between two
classes each following a Gaussian distribution,  but not for 
two classes forming non-overlapping concentric rings one inside the
other.  It is our hope that more measures used in combination will provide 
a more complete picture about class separation, which determines the
difficulty of classification. 

\clearpage
\section{Datasets for Validating Complexity Measures}

We evaluated the effectiveness of the complexity measures with
two collections of classification problems.
The first collection includes all pairwise discrimination problems 
from 14 datasets in the UC-Irvine Machine Learning Depository
\cite{kn:blake98}.  The datasets are those that contain at least 500
points with no missing values: {\em abalone, car, german, kr-vs-kp,
letter, lrs, nursery, pima, segmentation, splice, tic-tac-toe,
vehicle, wdbc, and yeast}.   Categorical features in some datasets
are numerically coded.  There are altogether 844 two-class
discrimination problems,  with training set sizes varying from 2
to 4648,  and feature space dimensionality varying from 8 to 480.
Using the linear programming procedure by Smith \cite{kn:smith68}
(as given in the description of the L1 measure in
Table \addtocounter{table}{-1}\thetable\addtocounter{table}{+1}),
452 out of the 844 problems are found to be linearly separable.   The
class boundary in each of these 
problems, as far as the training set is concerned, can be described
entirely by the weight vector of the separating hyperplane,  so by
Kolmogorov's notion these are simple problems.  Thus a valid
complexity measure should place these problems at one end of its scale.

The second collection consists of 100 artificial two-class problems each having
1000 points per class.  Problem 1 has one feature dimension, problem 2
has two, so forth and the last problem contains 100 features.
Each feature is a uniformly distributed pseudorandom number in
$[0,1]$.  The points are randomly labeled, with equal probability, as
one of two classes.  Therefore, these are intrinsically complex
problems, and they are expected to locate at the other end of any
complexity scale.

We studied the complexity measures on the distribution of these three
groups of problems, namely, (1) UCI linearly separable, (2) UCI
linearly nonseparable, and (3) random labelings.  A single measure
is considered useful for describing problem complexity if the three
groups of problems are separable on its scale,  and a set of measures are
considered useful if the groups of problems are separable in the space
spanned by the set.

\section{Key Observations}

The distribution of the three groups of classification problems
in this 12-dimensional complexity space displays many interesting
characteristics. A detailed description of the observations in this
study can be found in \cite{kn:ho02b}.  Here we summarize the main
findings. 

\subsection{Continuum of problem locations in complexity space}

The first remarkable observation in this study is that the
datasets fall on a continuum of positions
along many dimensions of the complexity space.   Even though there
have been no special selection criteria imposed on these naturally
arising datasets, we find that the problems cover a large range of
values in almost all the chosen complexity scales.  This reminds us of the
challenges in the practice of pattern recognition:  to pursue a good
match of methods to problems,  we must make sure that the classifier
methodology we choose is robust to variations in these problem
characteristics,  or we must understand the nature of the dependence 
of classifier behavior on such variations.  Without accomplishing
either,  applications of classifiers to problems are nothing but a
blind match,  and there is little hope of ensuring highest success. 

A more encouraging observation is that many of the real-world (UCI)
datasets are located far away from the random labelings,  suggesting that
these practical problems do indeed contain some intrinsic,
learnable structure.

Interestingly,  there is substantial spread among the random labelings
of different dimensionality.  While there is no obvious 
explanation for how dimensionality affects their intrinsic
difficulties,  closer examination of the differences suggests that 
this is more an effect of differences in apparent complexity due to
different sampling densities,  since these datasets all have the same
size while the volume of the space increases exponentially with dimensionality.

\subsection{Effectiveness of individual measures in separating problems
of known levels of difficulty}

The concentrations of the three groups of datasets (UCI linearly
separable, UCI linearly nonseparable, and random labelings) in
different regions in the complexity space suggest
that many of the measures can reveal their differences.   As a
stand-alone scale of complexity,  several measures (F1,F2,F3,L2,L3) are
especially effective in separating at least two of the three groups,
with the easiest set (UCI linearly separable) 
and the most difficult set (random labelings) occupying two opposite
ends of the scale.   However,  none of the measures can completely
separate the three groups with no overlaps.   Some measures, such as
N4 and T2, are especially weak when used in isolation.

\subsection{Distorted nearest neighbor error rates of sparse datasets}

The nearest-neighbor related measures (N1,N2,N3) have almost the same
discriminating power for the three groups, except for a few peculiar
cases where the training set consists of only 2 or 3 points.   For
those extremely sparse datasets,  although the class boundary (for the
training set) is linear, 
the nearest neighbors are almost always in a wrong class,  thus 
the nearest-neighbor error rate becomes very high.   This is an artifact
of the leave-one-out estimate.   However,  it also suggests 
that a single error rate, even that of a simple and
well-understood classifier, may tell a distorted story about the data
complexity.

\subsection{Pairwise Correlations between complexity measures}

Bivariate plots of the distributions show that some pairs of measures,
such as L2 and L3, or N1 and N3, are strongly correlated, while little
correlation is seen between many other pairs
(Table \addtocounter{table}{+1}\thetable\addtocounter{table}{-1}).
The existence of many uncorrelated pairs suggests that there are more
than one independent factors affecting a problem's complexity.

An examination of the correlation between L2 (linear classifier error rate)
and N3 (nearest neighbor error rate) and between each of these two
measures and others suggests that these error rates are not perfectly
correlated,  nor are they always predictable by an arbitrary  
measure.  This reconfirms the risk of relying on simple classifier
error rates for complexity measurement.   These two classifiers,
operating on very different principles (linearity versus proximity),
have difficulties caused by different characteristics of a problem
(Figure \addtocounter{figure}{1}\thefigure\addtocounter{figure}{-1}).

Some measures, while on their own are very weak in separating all 
three groups of problems,  can reveal the group differences when used
in combination with other measures
(Figure \addtocounter{figure}{2}\thefigure\addtocounter{figure}{-2}).
This demonstrates the importance of examining the multiple aspects of a
problem's complexity jointly.

The measure T1, while on its own being a strong separator of the three groups,
characterizes a very different aspect of complexity from
others as evidenced by its weak correlation with others.  
Inspection of the plots involving T1 and others suggests that while the
shapes of the classes can vary a lot across different problems,  it is
less relevant to classification accuracy than the shapes of the class
boundaries.  

\begin{table}[htbp]
\center
\begin{tabular}{|l|r r r r r r r r r r r r|}\hline
 & F1 & F2 & F3 & L1 & L2 & L3 & N1 & N2 & N3 & N4 & T1 & T2 \\ \hline
F1 & 1.00 & -0.02 & 0.06 & -0.01 & -0.02 & -0.02 & 0.07 & 0.01 & 0.14 & -0.02 & 0.03 & -0.03 \\
F2 & & 1.00 & -0.53 & 0.07 & 0.91 & 0.91 & 0.69 & 0.71 & 0.61 & 0.17 & 0.28 & 0.19 \\
F3 & & & 1.00 & -0.24 & -0.65 & -0.62 & -0.39 & -0.69 & -0.29 & -0.40 & -0.68 & -0.28 \\
L1 & & & & 1.00 & 0.33 & 0.32 & 0.37 & 0.37 & 0.28 & 0.53 & 0.18 & -0.10 \\
L2 & & & & & 1.00 & 1.00 & 0.78 & 0.81 & 0.67 & 0.47 & 0.37 & 0.16 \\
L3 & & & & & & 1.00 & 0.78 & 0.81 & 0.67 & 0.46 & 0.35 & 0.16 \\
N1 & & & & & & & 1.00 & 0.76 & 0.96 & 0.49 & 0.39 & -0.02 \\
N2 & & & & & & & & 1.00 & 0.68 & 0.51 & 0.55 & 0.12 \\
N3 & & & & & & & & & 1.00 & 0.38 & 0.38 & -0.04 \\
N4 & & & & & & & & & & 1.00 & 0.28 & 0.30 \\
T1 & & & & & & & & & & & 1.00 & 0.17 \\
T2 & & & & & & & & & & & & 1.00 \\
\hline
\end{tabular}
\caption{Correlation coefficients between each pair of measures.}
\end{table}

\begin{figure}[ht]
\centering
\begin{tabular}{cc}
\psfig{file=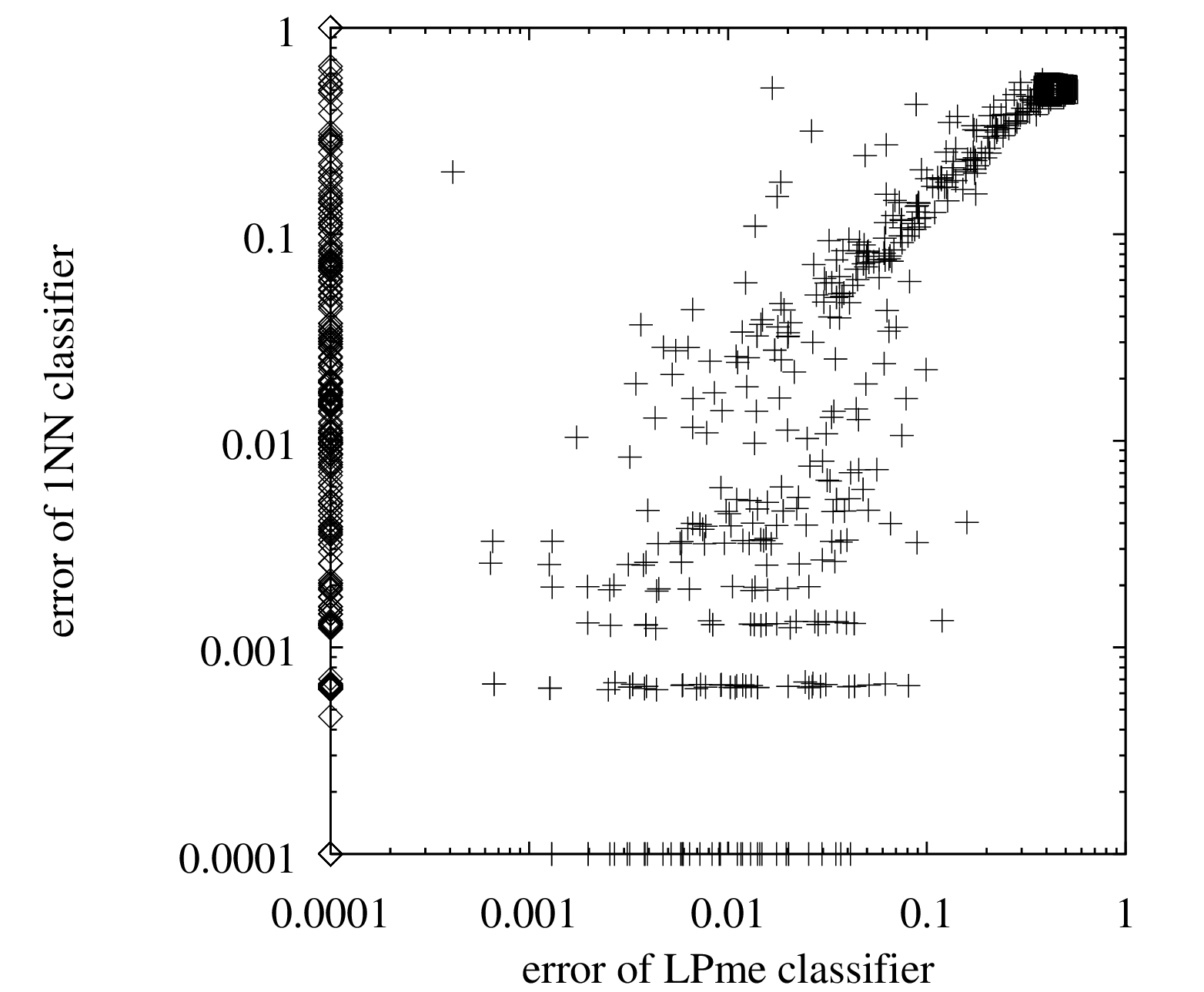,width=2in,clip=}&
\hspace{0.25in}
\psfig{file=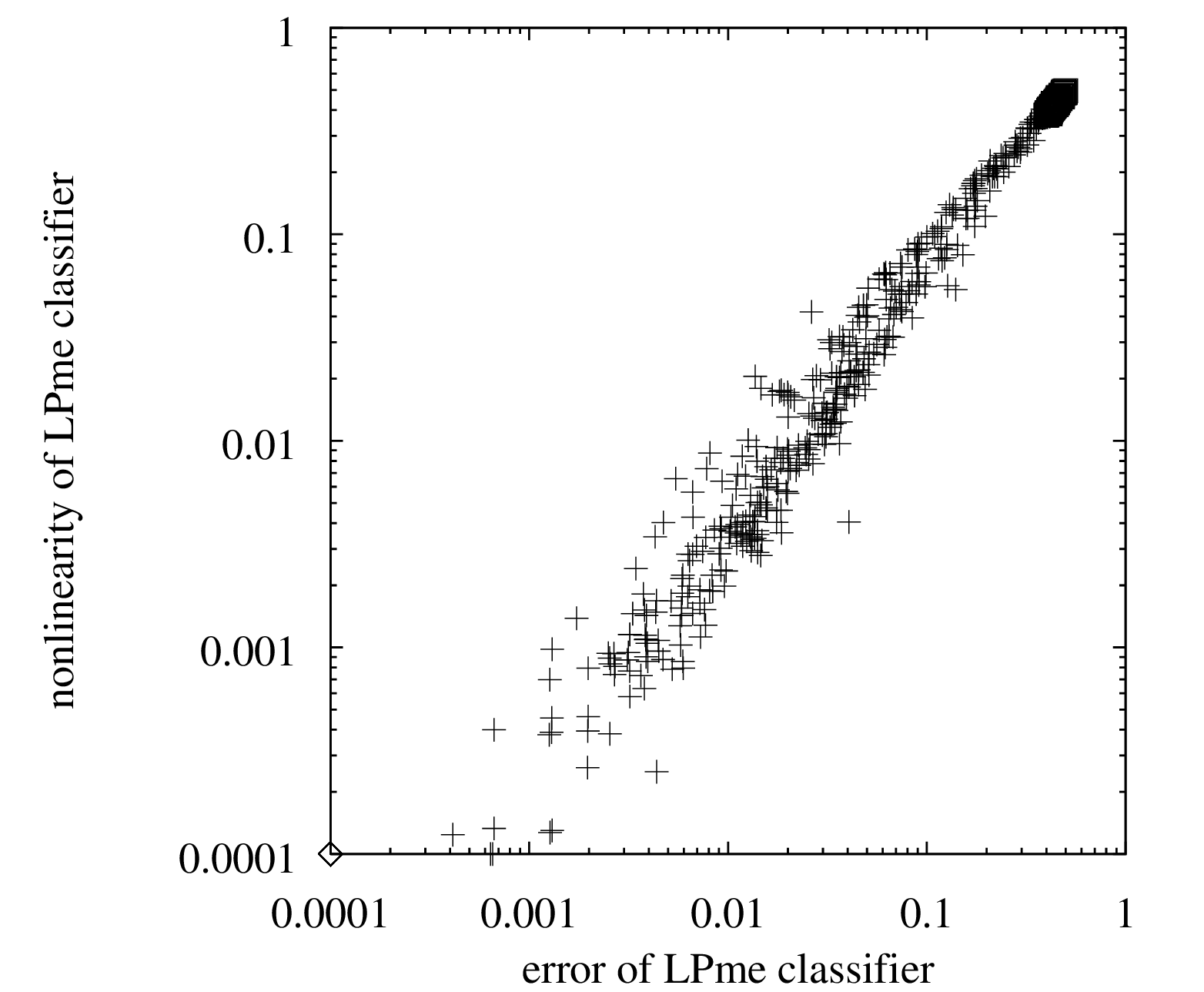,width=2in,clip=}\\
\psfig{file=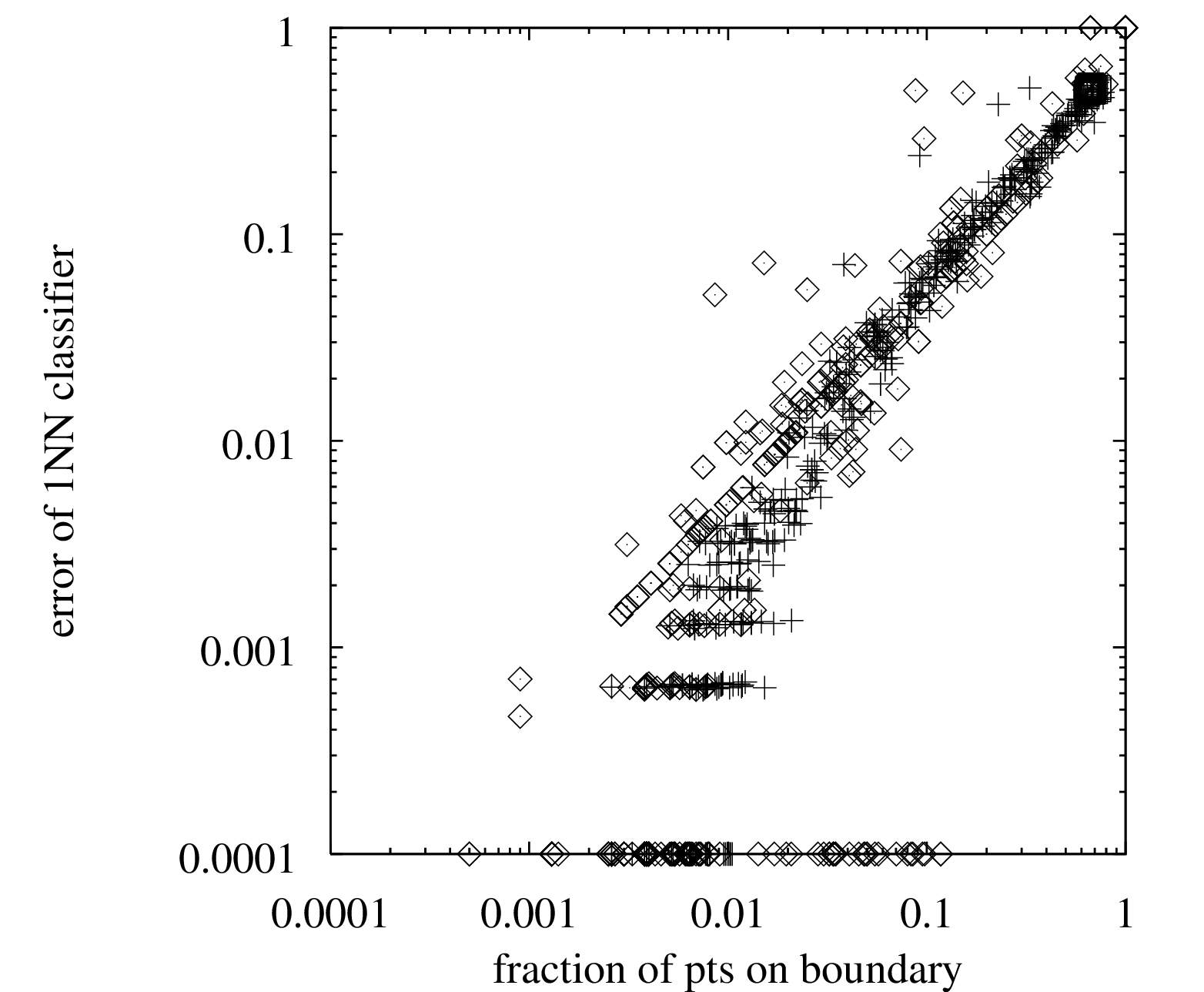,width=2in,clip=}&
\hspace{0.25in}
\psfig{file=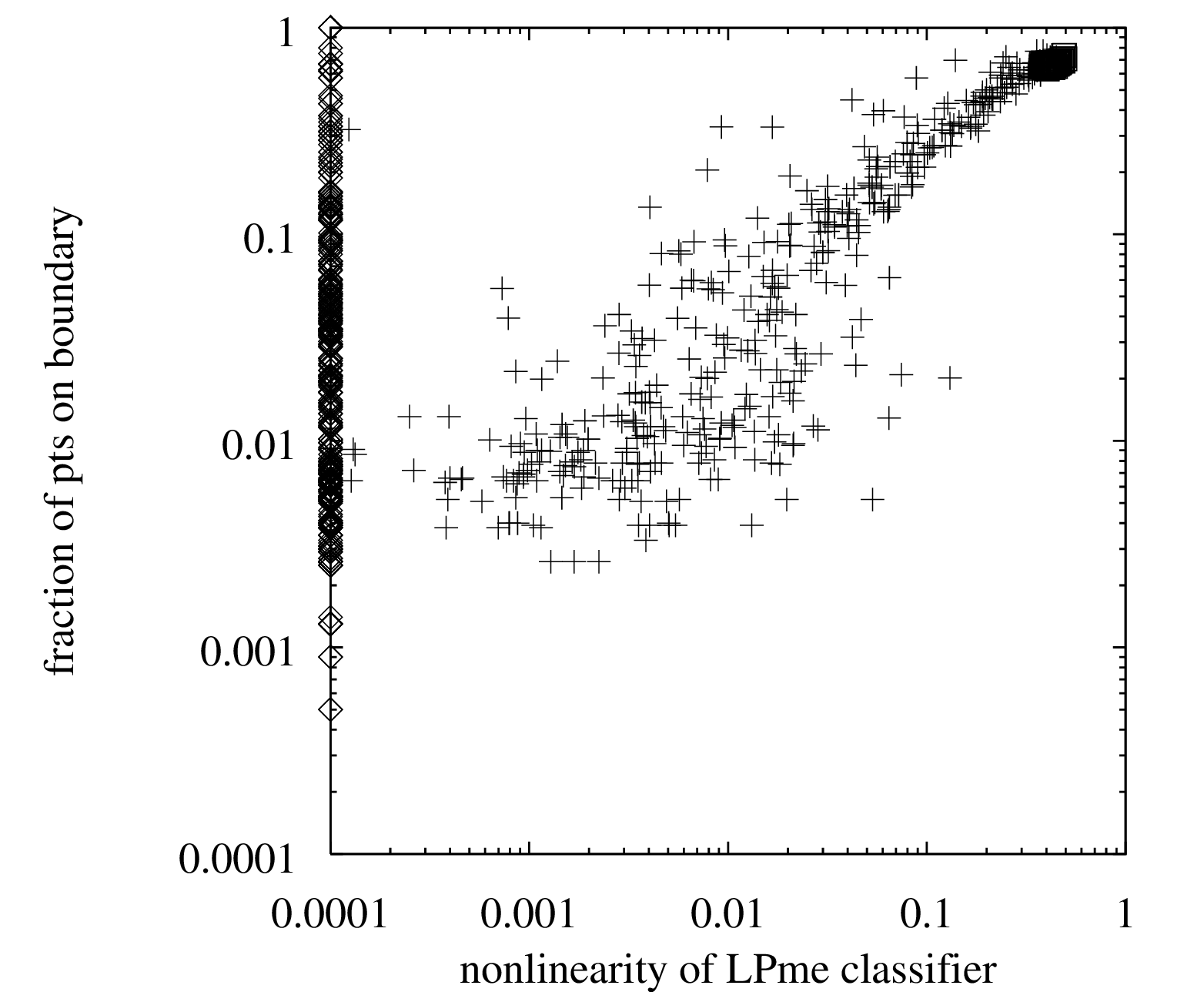,width=2in,clip=}\\
\end{tabular}
\caption{Error rates of linear and nearest-neighbor classifiers, and
the measures they are most strongly correlated with.
($\diamond$: UCI linearly separable problems; $+$: UCI linearly
nonseparable problems; $\Box$: random labelings)}
\end{figure}

\begin{figure}[ht]
\centering
\begin{tabular}{cc}
\psfig{file=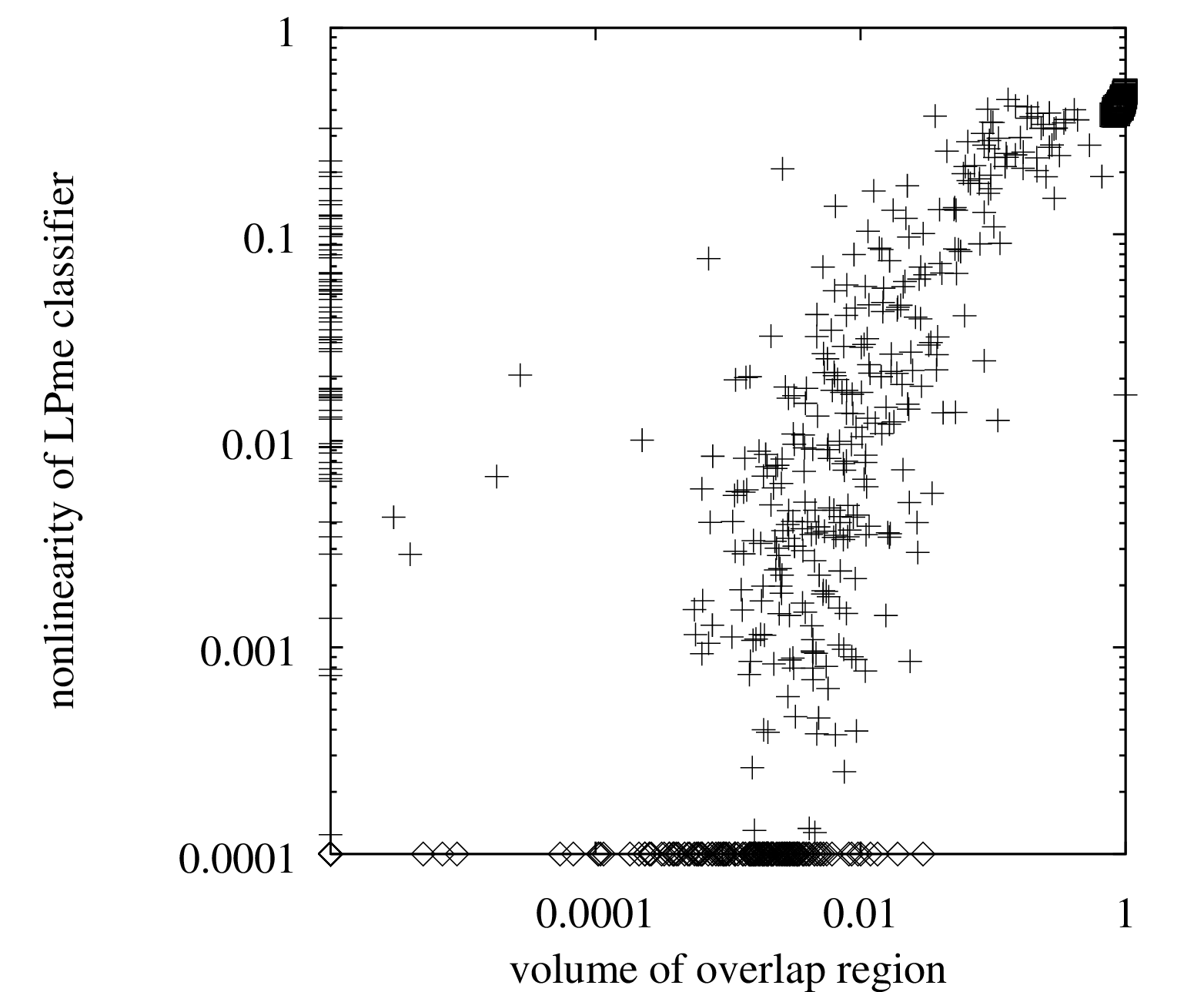,width=2in,clip=}&
\hspace{0.25in}
\psfig{file=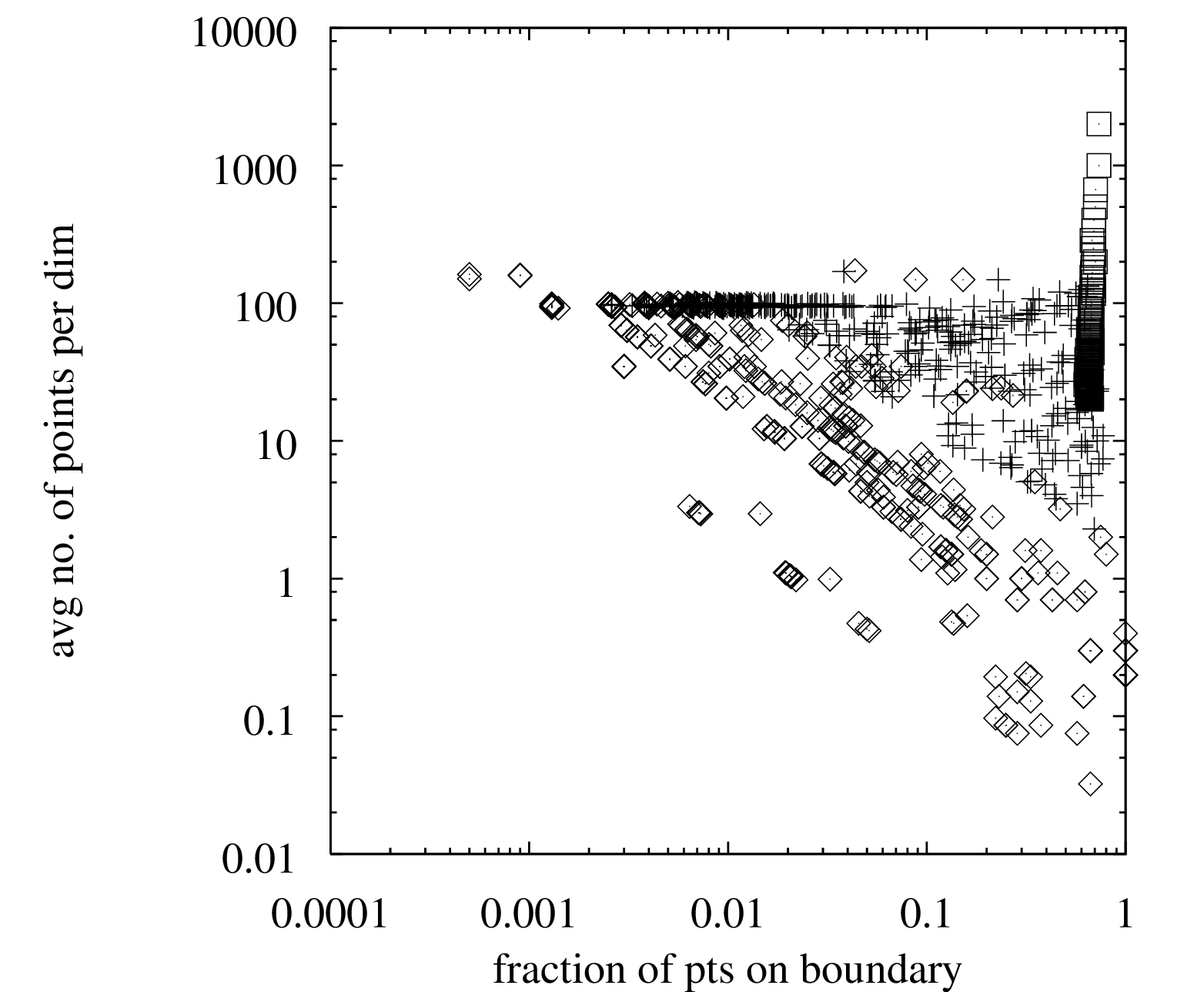,width=2in,clip=}\\
\end{tabular}
\caption{Groups of problems that overlap heavily on an individual
complexity scale may show clear separation in the interaction of the effects.
($\diamond$: UCI linearly separable problems; $+$: UCI linearly
nonseparable problems; $\Box$: random labelings)}
\end{figure}

\subsection{Principal components of the complexity space}

A principal component analysis using the distribution of the problems
in the 12-dimensional space shows that there are six significant
components each explaining more than 5\% of the variance.
Among these, the first component explains over 50\% of the variance,
and comprises even contributions from F2,L2,L3,N1,N2, and N3.  It is a
combination of effects of linearity of class boundaries and proximity
between opposite class neighbors.  The next 3 components explain
12\%, 11\% and 9\% of the variance respectively,  and can be
interpreted as (PC2:) a balance of within-class and between-class
scatter,  (PC3:) the concentration and orientation of class overlaps, 
and (PC4:) within-class scatter.   For a more detailed discussion of
these components, as well as for the trajectory traced in the PC
projection by an example series of problems with controlled class
separation,  we again refer readers to \cite{kn:ho02b}.

\section{Studies of Problems and Classifiers Using Complexity Measures}

A complexity measurement space like this has many potentially
interesting uses.  For a particular application domain,  the scales of
complexity can help determine the existence of any learnable
structure, which can be used to set expectations on automatic learning
algorithms.   They can also be used to determine if a particular
dataset is suitable for evaluating different learning algorithms.

The measures can be used to compare different problem formulations,
including class definitions, choice of features, and potential feature
transformations.  They can be used to guide the selection of
classifiers and classifier combination schemes,  or control the process of
classifier training.  A use of these measures for comparing two
methods for decision forest construction is reported in \cite{kn:ho02c}.  

Regions occupied by datasets on which classifiers display homogeneous
performances can be used to outline the domain of competences of
those classifiers,  with the expectation that performances
on new datasets falling in the same region can be predicted
accordingly.  Regions where no classifiers can do well may be
characterized in detail by the complexity measures,  which could lead
to new classifier designs covering those blind spots.  
In \cite{kn:ester03} we report a study of the domain of competence of
XCS, a genetic algorithm based classifier. 

One may wish to study the distribution of all classification problems
in this space.  An empirical approach will be to seek a representation
of the distribution by a much larger collection of practical problems.
A theoretical approach will be more challenging;  it involves
reasoning about regions in this space that are possible or impossible
for any dataset to occupy.   The identification of such regions will
require a better understanding of constraints in high-dimensional data
geometry and topology.   The intrinsic dimensionality of
the problem distribution will give more conclusive evidence on how many
independent factors contribute to a problem's difficulty.

\section{Conclusions}

We describe some early investigation into the complexity of a
classification problem,  with emphasis on the geometrical characteristics
that can be measured directly from a training set.  We took some
well known measures from the pattern recognition literature,  and
studied their descriptive power using a collection of problems of
known levels of difficulty.  We found some interesting spread among the
different types of problems,  and evidence of existence of independent
factors affecting a problem's difficulty. 
We believe that such descriptors of complexity are useful for
identifying and comparing different classes of problems,
characterizing the domain of competence of classifier or ensemble
methods, and in many ways guiding the development of a solution to a
pattern recognition problem.

These are our first steps towards developing elements of a language with
which we can talk more precisely about properties of high dimensional
datasets, especially those aspects affecting classifier performances.
We believe this is necessary for classification research to advance
beyond the current plateau.   Finally, we believe that such abstract
studies are best coupled with tools for interactive visualization of
a dataset,  so that an intuitive understanding may be obtained on
how the complexity arises from a particular problem \cite{kn:ho02d}.

\end{document}